\documentclass[letterpaper, 10 pt, conference]{ieeeconf}  

\IEEEoverridecommandlockouts                              

\usepackage{courier}        
\usepackage{type1cm}        
\usepackage{placeins}
\usepackage{amsmath}

\DeclareMathOperator*{\argmin}{arg\,min}
\usepackage{amssymb}
\usepackage{makeidx}         
\usepackage{graphicx}        
\usepackage{subcaption}\usepackage[font=footnotesize]{caption}
\newlength{\twosubht}
\newsavebox{\twosubbox}
\usepackage{multicol}        
\usepackage[bottom]{footmisc}
\usepackage{enumerate}
\let\labelindent\relax

\usepackage{enumitem}
\usepackage[all]{xy}
\usepackage{cite}
\usepackage[ruled]{algorithm}
\usepackage[noend]{algpseudocode}
\usepackage[usenames, dvipsnames]{color}
\usepackage{bm}
\usepackage{mathrsfs}
\usepackage{graphicx}
\usepackage{epsfig}
\usepackage{pdfpages}
\usepackage{wrapfig}
\usepackage{lipsum}
\usepackage{times}
\usepackage{setspace}
\usepackage{cancel}
\usepackage{wrapfig}
\usepackage[rightcaption]{sidecap}
\usepackage{adjustbox}
\usepackage{xspace}
\usepackage{float}
\usepackage{mathtools}
\usepackage{breqn}
\usepackage{nicefrac}

\usepackage[pagebackref=false,breaklinks=true,letterpaper=true,colorlinks,bookmarks=false]{hyperref}

\hypersetup{%
  colorlinks = true,
  pdfauthor  = {Rebecca H. Jiang},
  pdftitle   = {Parallel-Jaw Gripper and Grasp Co-Optimization for Sets of Planar Objects},
  urlcolor = black}

\usepackage[short]{optidef}
\usepackage{titlesec}
\titlespacing*{\section}
{0pt}{1ex plus 0ex minus .2ex}{1ex plus 0ex}
\titlespacing*{\subsection}
{0pt}{1ex plus 0ex minus .2ex}{1ex plus 0ex}
\titlespacing*{\appendix}
{0pt}{1ex plus 0ex minus .2ex}{1ex plus 0ex}

\usepackage[nameinlink]{cleveref}
\crefname{equation}{Eq.}{Eq.}
\crefname{figure}{Fig.}{Figs.} 
\crefname{section}{Sec.}{Secs.}
\crefname{subsection}{Sec.}{Secs.} 
\crefname{table}{Table}{Tables} 
\crefname{appendix}{Appendix}{Appendices}
\crefname{problem}{Prob.}{Probs.}
\creflabelformat{problem}{(#1)#2#3}
\crefname{inequality}{Ineq.}{Ineqs.}
\creflabelformat{inequality}{(#1)#2#3}

\def\equationautorefname~#1\null{(#1)\null}

\newcounter{problem}

\newcounter{inequality}

\title{\LARGE \bf
Parallel-Jaw Gripper and Grasp Co-Optimization for Sets of Planar Objects
}

\author{Rebecca H. Jiang$^{1}$, Neel Doshi$^{2}$, Ravi Gondhalekar$^{3}$, Alberto Rodriguez$^{4}$
\vspace{-3ex}
\thanks{$^{1}$Rebecca H. Jiang is with the Department of Aeronautics and Astronautics, Massachusetts Institute of Technology and is a Draper Scholar at The Charles Stark Draper Laboratory, Inc.
        {\tt\small rhjiang@mit.edu}}%
\thanks{$^{2}$Neel Doshi is with Amazon Robotics R\&D.  This research was conducted prior to Neel joining Amazon
        {\tt\small ndd@amazon.com}}%
\thanks{$^{3}$Ravi Gondhalekar is with The Charles Stark Draper Laboratory, Inc.
        {\tt\small rgondhalekar@draper.com}}
\thanks{$^{4}$Alberto Rodriguez is with the Department of Mechanical Engineering, Massachusetts Institute of Technology
        {\tt\small albertor@mit.edu}}%
}

\begin{document}

\linepenalty = 1000

\maketitle
\thispagestyle{empty}
\pagestyle{empty}

\begin{abstract}
We propose a framework for optimizing a planar parallel-jaw gripper for use with multiple objects.  While optimizing general-purpose grippers and contact locations for grasps are both well studied, co-optimizing grasps and the gripper geometry to execute them receives less attention.  As such, our framework synthesizes grippers optimized to stably grasp sets of polygonal objects. Given a fixed number of contacts and their assignments to object faces and gripper jaws, our framework optimizes contact locations along these faces, gripper pose for each grasp, and gripper shape.  Our key insights are to pose shape and contact constraints in frames fixed to the gripper jaws, and to leverage the linearity of constraints in our grasp stability and gripper shape models via an augmented Lagrangian formulation. Together, these enable a tractable nonlinear program implementation.  We apply our method to several examples.  The first illustrative problem shows the discovery of a geometrically simple solution where possible. In another, space is constrained, forcing multiple objects to be contacted by the same features as each other. Finally a toolset-grasping example shows that our framework applies to complex, real-world objects. We provide a physical experiment of the toolset grasps.
\end{abstract}

\section{Introduction}\label{sec:introduction}
\begin{figure*}[t]

\sbox\twosubbox{%
  \resizebox{\dimexpr \linewidth-1em}{!}{%
    \includegraphics[height=1cm]{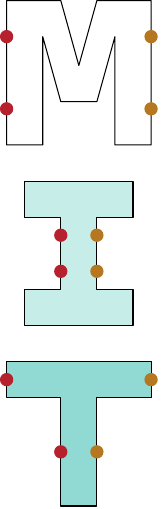}%
    \includegraphics[height=1cm]{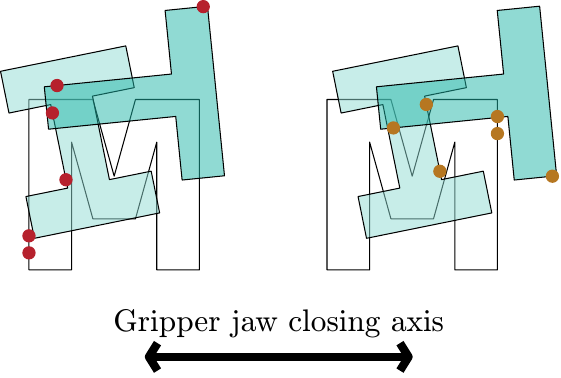}%
    \includegraphics[height=1cm]{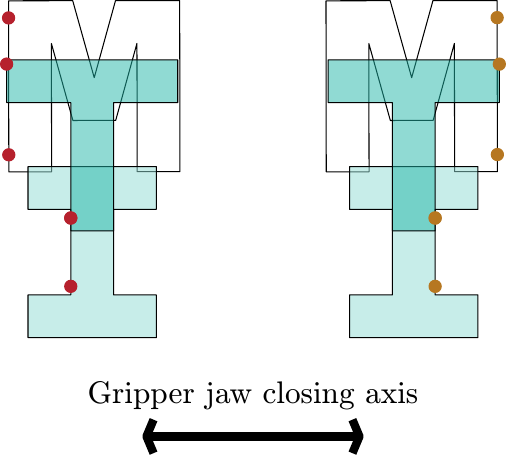}%
    \includegraphics[height=1cm]{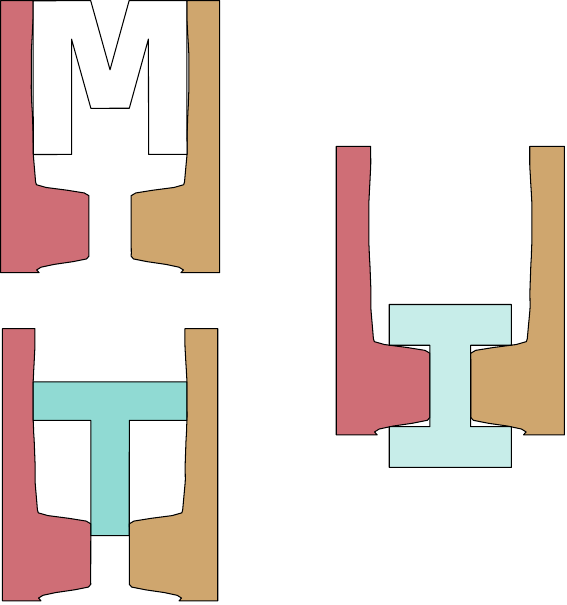}%
  }%
}
\setlength{\twosubht}{\ht\twosubbox}
\centering
\subcaptionbox{\label{fig:intro1}}{%
  \includegraphics[height=0.95\twosubht]{figures/MITLettersDef.pdf}%
}\quad
\subcaptionbox{\label{fig:intro2}}{%
  \includegraphics[height=0.95\twosubht]{figures/MITLettersBadGuess.pdf}%
}
\quad
\subcaptionbox{\label{fig:intro3}}{%
  \includegraphics[height=0.95\twosubht]{figures/MITLettersGoodGuess.pdf}%
}\quad
\subcaptionbox{\label{fig:intro4}}{%
  \includegraphics[height=0.95\twosubht]{figures/MITLettersFinalGrasp.pdf}%
}
\caption{(a) Task input information: three letter-shaped objects and contact assignments to object edges and gripper jaws.  Here and throughout this work, red contact points and gripper shapes correspond to the left gripper, and orange to the right. (b) and (c) Shapes and contact points in the gripper frames (left image in $G_L$ and right in $G_R$) for an infeasible and optimized (respectively) candidate solution for configuration variables $\mathbf{z}$. The gripper jaw closing axis is horizontal in these gripper frames. (d) A gripper solution corresponding to the optimized configuration variables from (c).  }
\label{fig:presolve}
\vspace{-18pt}
\end{figure*}

While many works study grasp optimization \cite{nguyen1988constructing,ferrari1992planning,roa2009computation,hang2014combinatorial} and gripper design, gripper shape and gripper-object contact are rarely reasoned about together. As a result, these optimized grasps are only realized by dexterous manipulators \cite{rosales2012synthesis,miller2004graspit,brahmbhatt2019contactgrasp}, which incur substantial mechanical and control complexity.  We propose a framework for optimizing the contact surfaces of parallel-jaw grippers subject to a grasp stability model.  For a set of input objects like in \cref{fig:intro1}, our framework optimizes a single parallel-jaw gripper that can stably grasp each object, like in \cref{fig:intro4}.

Designing grippers involves considering shape in two contexts: a) how does the gripper contact the objects?, and b) what form does the gripper surface take in between contact points?  While the former alone determines grasp quality, both determine geometric compatibility with the target objects.  For example, Kodnongbua et al. \cite{kodnongbua2022computational} first select contact points for passive grippers, and subsequently find collision-free grippers to meet those contacts.  In our previous work \cite{jiang2022shape}, we co-optimize shape and motion of rigid effectors to contact moving objects, constraining all points on the contact surface to be collision-free at all times.  However, in this work the goal is to optimize parallel-jaw grippers; in addition to the complications of adding an actuated degree of freedom, simultaneously optimizing over contact locations and many gripper shape parameters creates a large, nonconvex design space.  Further complicating this problem is the need to consider geometric compatibility and stability of grasps of multiple target objects across different gripper poses, when these poses are also decision variables. 

Our key contribution is to pose gripper optimization as a tractable nonlinear program (NLP) (\cref{sec:problem}).
We consider for each candidate gripper the stability of the resulting grasps (\cref{sec:graspStability}), and the properties of the gripper surfaces modeled expressively as piecewise polynomials, on which contact and non-penetration constraints are enforced (\cref{sec:gripperShape}).  We pose the problem tractably by leveraging its underlying structure.  A candidate set of values for \textit{configuration variables} -- contact locations and gripper configurations for each grasp -- allows the objects and contacts to be transformed into \textit{gripper frames} fixed to the gripper jaws, as visualized in \cref{fig:intro2} and \cref{fig:intro3}.  This representation allows us to pose grasp stability and shape considerations as convex quadratic programs (QPs) whose parameter matrices are functions of the configuration variables.  An augmented Lagrangian formulation allows us to optimize over all problem variables jointly while still leveraging QP solvers to solve the QPs to global optimality (\cref{sec:optimization}). We apply our framework to three examples and show a real-world demonstration in \cref{sec:results}.  We discuss computational cost of our framework in \cref{sec:params}.

\section{Related work}\label{sec:relatedWorks}
In this section we discuss works in the well-studied problem of optimizing general-purpose grippers, as well as the less-studied problem of optimizing task-specific grippers.

\subsection{General-purpose gripper optimization}
Several works search over small sets of geometric parameters, optimizing for desirable gripper behavior \cite{ciocarlie2013kinetic,liu2022simulation}, simple metrics for grasp stability \cite{dong2018geometric,russo2017design,lanni2009optimization}, and force transmission \cite{krenich2004multicriteria,saravanan2009evolutionary,hassan2017modeling}.  Beyond these considerations, Elangovan et al.\ \cite{elangovan2021improving} maximize manipulation workspace, and Yako et al. \cite{yako2022designing} use a potential energy map to understand grasping behavior without simulation. 
These approaches show success in deciding parameters, but have limited expressivity compared to higher-dimensional design spaces like in our work. 

Other works use topology optimization to maximize mechanical advantage \cite{liu2019topology,liu2020optimal} and tip deflection \cite{zhang2017design,zhang2018design,chen2018topology,wang2020topology} of soft grippers. These formulations are highly expressive, but do not leverage object or task information, making them unsuitable for optimizing grasps for sets of objects.  

\subsection{Task-specific gripper optimization}

Specifying a target object, or set of objects, enables grasp simulation. Wolniakowski et al. \cite{wolniakowski2015task} optimize dimensions of a simple gripper, maximizing simulated grasp success and robustness to pose perturbations, among other objectives. Using a learning-based approach, Ha et al.\ \cite{Ha2020fit2form} and Alet et al.\ \cite{aletrobotic} encourage robustness by evaluating via simulating with multiple initial object poses.  Like in our approach, Alet et al.\ design grippers for sets of objects.  These two works show good tolerance to uncertainty, but do not fully leverage model information or the powerful tools of grasp stability analysis to design grippers with highly tailored geometry. 
 
Schwartz et al.~\cite{schwartz2017designing} leverage object models, designing grippers based on ``imprints,'' or the negatives of the object shapes.  Honarpardaz et al.~\cite{honarpardaz2017fast} similarly imitate the object shape with a contact surface.  While neither of these works reasons explicitly about grasp quality, Brown and Brost \cite{brown19993} design grippers for form closure grasps. Like the present work, they score grasps via a point-contact model and reason about non-penetration during jaw
closure. However, they fix object orientation within
the jaws, omitting an important freedom that we include.  
 Finally, Kodnongbua et al.~\cite{kodnongbua2022computational} design passive grippers.  They first rank sets of contact locations, then use topology optimization to design a collision-free gripper geometry and approach path.  While their results show impressive use of the design space and reliable performance, they optimize contact locations and gripper shape in separate stages, restricting generality. Furthermore, in these works that leverage object models, tailoring grippers to objects comes at the expense of multi-purposing: each resulting gripper is compatible with only one target object. In contrast, our framework designs a gripper for a
 set of target objects.

\section{The gripper optimization problem}\label{sec:problem}
In this section, we discuss problem formulation, representation, and assessing the feasibility of a candidate solution.

\subsection{Problem formulation and notation}\label{sec:formulation}
The following items are required problem inputs:
\begin{itemize}[leftmargin=*,wide]
    \item Descriptions of polygonal objects, $\Psi^W[k]$, in the world frame ($W$), $k=0,...,N_o$.
    \item Object edge assignments for contacts: 
    A set of contact indices $E_e:=\{i|$ contact $i$ is on edge $e\}$ for each object edge $e$. Object $k$ has $N_c[k]$ contacts.
    
    \item Gripper jaw assignments for contacts: a set of contact indices $C_j:=\{i|$ contact $i$ belongs to jaw $j\}$ for each jaw $j\in \{L,R\}$. 
\end{itemize}

If a solution is found, the framework outputs:
\begin{itemize}[leftmargin=*,wide]
\item Gripper configurations: position $\mathbf{p}^W_{G}[k]$, orientation $\theta_G[k]$, and jaw opening $\gamma[k]$, of the gripper ($G$), for grasping each object. We abbreviate $\mathcal{P} := \{\mathbf{p}^W_{G}[0],..., \mathbf{p}^W_{G}[N_o]\}$, $\Theta := \{\theta_G[0],...,\theta_G[N_o]\}$, $\Gamma := \{\gamma[0],...,\gamma[N_o]\}$.
\item Positions of contact points along the edges of the polygonal object: $d_{i}[k]$, the $i$th contact's distance from an edge vertex, normalized by the edge length.  We abbreviate $\mathbf{d}[k]:=[d_{0}[k],...,d_{N_c[k]-1}[k]]^\top$ for each object, and  $\mathcal{D}:=\{ \mathbf{d}[0],...,\mathbf{d}[N_o]\}$.
\item Gripper surface parameterization: positions $v_L[i_y]$, $v_R[i_y]$ and slopes $m_L[i_y]$, $m_R[i_y]$ of the left and right gripper surfaces (in left and right gripper frames $G_L$ and $G_R$) along a grid of vertical coordinates $\mathbf{y}:=[y[0],...,y[N_{y}]]^\top$, for piecewise cubic Hermite interpolation. We group all positions as $\mathbf{V}:=[v_L[0],...,v_L[N_y],v_R[0],...,v_R[N_y]]^\top$, and all slopes $\mathbf{M}:=[m_L[0],...,m_L[N_y],m_R[0],...,m_R[N_y]]^\top$.
\end{itemize}
In addition to these inputs and outputs, the optimization problem is parameterized by: shape cost component weights $w_s$, $w_p$, curvature cost weight parameter $\sigma$, shape program constraints weight $\rho_S$, and penalty parameter increase rate $\phi$.
\subsection{Approach}\label{sec:approach}

The challenge in optimizing for the output variables in \cref{sec:formulation} is that this space is large and nonconvex.  In particular, to yield an expressive shape representation, the grid size $N_y$ must be large. Naively formulating an NLP to directly optimize over all these variables is impractical.  Instead, we pose the problem as a convex QP whose parameter matrices are functions of \textit{configuration variables} $\mathbf{z}:=\{\Theta,\Gamma,\mathcal{P},\mathcal{D}\}$.  While we still need to use an NLP solver for $\mathbf{z}$, we solve a convex QP for the remaining variables (the much larger set).

For a candidate $\mathbf{z}$, we represent the objects and contact points in frames fixed to the left and right gripper surfaces, $G_L$, $G_R$, defined such that the gripper's axis of jaw actuation is horizontal. These representations allow us to pose considerations on \textit{grasp stability} and \textit{gripper shape}. \cref{fig:intro2,fig:intro3} show these gripper-frame representations for an infeasible and an optimized $\mathbf{z}$, respectively, corresponding to problem input in \cref{fig:intro1}. \cref{fig:intro4} shows a gripper solution corresponding to the optimized configuration in \cref{fig:intro3}.

\textbf{Grasp stability.}
In \cref{fig:intro2}, the contact points on the letter \textit{M} from either gripper jaw are vertically misaligned -- the gripper jaws squeezing along the horizontal axis would create a net torque on the object, preventing the static equilibrium necessary for a grasp. The letter \textit{T} faces contacted are nearly parallel to the jaw axis, poorly situated to transmit normal forces to preload the grasp through jaw squeezing.  In contrast, in \cref{fig:intro3}, the candidate $\mathbf{z}$ results in good grasps, where 
contacts oppose each other and contact normals are aligned with the jaw axis.  Grasp stability feasibility and a quality metric, as functions of $\mathbf{z}$, are formalized in \cref{sec:graspStability}.

\textbf{Gripper shape.}
In \cref{fig:intro2}, contact points on the letter \textit{I} lie inside the \textit{M} and \textit{T} objects.  Parts of the gripper surface that contact the \textit{I} would necessarily penetrate the \textit{M} and \textit{T} when those objects are grasped; designing a gripper shape for this candidate $\mathbf{z}$ is infeasible.  Gripper shape feasibility and a shape metric, as functions of $\mathbf{z}$, are formalized in \cref{sec:gripperShape}.

\section{Grasp stability}\label{sec:graspStability}
\begin{figure}[t]
    \centering
    \includegraphics[width=\columnwidth]{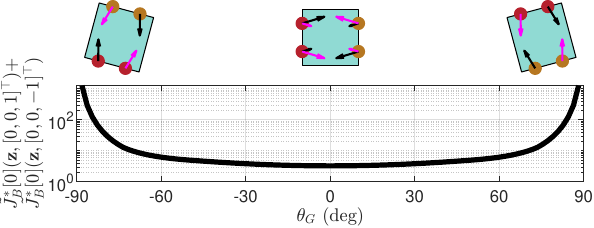}
    \caption{Grasp quality metric for square plotted as a function of grasping orientation.  Above: square plotted in gripper frame $G_L$ at $\theta_G=$ -75$^\circ$, 0$^\circ$, and 75$^\circ$. Jaw closing axis is horizontal. Contact force solutions plotted in black and magenta arrows for $\mathbf{w}=[0,0,1]$ and $\mathbf{w}=[0,0,-1]$, respectively.} 
    \label{fig:graspStability}
    \vspace{-2pt}
\end{figure}
In this section, we develop a convex QP to assess the stability of a candidate grasp, selecting candidates like the one shown in \cref{fig:intro3} and discouraging those like that in \cref{fig:intro2}. There exist many frameworks that check for grasp stability by searching for contact forces that satisfy linear static-equilibrium equations.  However, these methods find any statically feasible contact forces instead of considering what forces may actually arise as a result of actuated degrees of freedom (in this case, the squeezing action of the parallel jaws). To resolve this, we use two concepts from Haas-Hegar et al.~\cite{haas2018passive,haas2021grasp}: considering preload in the grasping model, and a compliance model for resolving static indeterminacy.

We consider grasp stability for each object individually.  For each object, the \textit{grasp matrix} $\mathcal{G}[k]$ and \textit{hand Jacobian} $\mathcal{J}[k]$ can be computed as functions of gripper orientation $\theta_G[k]$ and contact coordinates $\mathbf{d}[k]$.  In addition to contact forces $\mathbf{c}:=[c_{0,n},c_{0,t},...,c_{N_c[k]-1,n},c_{N_c[k]-1,t}]$ ($n$ and $t$ indicate normal and tangent components, respectively), we include optimization variables for ``virtual displacements'' of the object, $\mathbf{r}\in \mathrm{SE}(2)$ and gripper jaws, $\mathbf{q}\in \mathbb{R}^2$. Normal and tangential displacements of ``virtual springs'' at the contacts, $\delta_{i,n}$, $\delta_{i,t}$,  can be calculated by transforming $\mathbf{r}$ and $\mathbf{q}$ into the $i$th contact frame, $[\delta_{0,n},\delta_{0,t},...,\delta_{N_c[k]-1,n},\delta_{N_c[k]-1,t}]^\top := (\mathcal{G}[k](\theta_G[k],\mathbf{d}[k]))^\top\mathbf{r}[k]-\mathcal{J}[k](\theta_G[k],\mathbf{d}[k])\mathbf{q}[k]$.  The full program we use to assess grasp stability is given by \autoref{eq:graspStabilityW}:
\begin{subequations}
\begin{align}
&\tilde{J}_{B}^*[k](\mathbf{z},\mathbf{w})  :=  \min_{\mathbf{r},\mathbf{q},\mathbf{c}} \quad \tilde{J}_{B}[k](\mathbf{r},\mathbf{q})&&\\
&\textrm{such that} \nonumber\\
& \mathcal{G}[k](\theta_G[k],\mathbf{d}[k])\mathbf{c} + \mathbf{w}= \mathbf{0} &\label{eq:gwEquilibrium}\\
 &c_{i,n} = -\delta_{i,n}&\forall& i \label{eq:gwSpring}\\
  &\delta_{i,n}\leq 0 &\forall& i \label{eq:gwSpringNonneg}\\
&[1,0](\mathcal{J}[k](\theta_G[k],\mathbf{d}[k]))^\top\mathbf{c} \geq 0 & \label{eq:gwHandEquilibrium}\\
&-\mu c_{i,n}\leq c_{i,t}\leq \mu c_{i,n} &\forall& i,\label{eq:gFriction}
\end{align}\label{eq:graspStabilityW}
\end{subequations}%
\hspace{-5pt}where $\tilde{J}_{B}[k]$ is a grasp quality metric for object $k$ (defined later), $\mu$ is a coefficient of friction, and $\mathbf{w}$ is an external wrench on the object, consisting of two forces and a torque, $\mathbf{w}=[f_x,f_y,\tau]^\top$. Constraint \autoref{eq:gwEquilibrium} enforces object static equilibrium, balancing gripper-applied contact forces with the external wrench.  Constraint \autoref{eq:gwSpring} enforces the compliance model: contact normal forces are proportional to virtual spring normal displacements, and \autoref{eq:gwSpringNonneg} enforces that the springs only compress, equivalent to enforcing non-negativity of normal contact forces.  The LHS of \autoref{eq:gwHandEquilibrium} evaluates to the net horizontal forces on the gripper jaws from object contact. Enforcing non-negativity of these forces models equilibrium for a preloaded parallel-jaw grasp. Note that feasibility of \autoref{eq:graspStabilityW} is not affected by the scale of $\mathbf{w}$;  when $\mathbf{w}\rightarrow\alpha\mathbf{w}$ for scalar $\alpha>0$, force and displacement solutions simply scale by $\alpha$. If we instead equated horizontal gripper jaw forces to a set preload value in \autoref{eq:gwHandEquilibrium}, we would lose this scale invariance. This is reasonable to do if actual values of both the intended grasp preload and the applied wrench $\mathbf{w}$  are known.  Finally, constraint \autoref{eq:gFriction} imposes friction cone bounds.

Cost function $\tilde{J}_{B}[k]$ should encourage good grasps such as the one shown in the top-middle of \cref{fig:graspStability}. This grasp is good because the axis of jaw action is aligned with contact normals, and contacts on either side of the square are vertically aligned.  As the grasping angle magnitude grows, the horizontal gripper jaw axis becomes poorly situated to apply normal forces at the contacts, requiring large virtual displacements $\mathbf{r}$ and $\mathbf{q}$ to achieve contact forces. Solutions to \autoref{eq:graspStabilityW} in this square-grasping example have unbounded $\mathbf{r}$ and $\mathbf{q}$ values as $\theta_G\xrightarrow{}\pm 90^\circ$.  This motivates the grasp quality metric \cref{eq:graspQuality}:
\begin{equation}\label{eq:graspQuality}
    \tilde{J}_{B}[k](\textbf{r},\textbf{q}) = \frac{1}{2L[k]^2}\left(\mathbf{r}^\top\mathbf{r} + \mathbf{q}^\top\mathbf{q}\right)
\end{equation}
where we nondimensionalize by length $L[k]$, defined as the mean distance from the object origin to the vertices on either end of an edge containing a contact point.

As the contacts are frictional and the grasp is preloaded, grasps feasible under \autoref{eq:graspStabilityW}, even if analyzed only with $f_x=f_y=0$, can withstand nonzero $f_x$ and $f_y$.  As such, in lieu of particular knowledge about actual intended loading, it is reasonable to check grasp stability with respect to just two wrenches, $\mathbf{w} = [0,0,\pm1]^\top$, and sum the resulting $\tilde{J}^*_B$s.  \cref{fig:graspStability} shows the resulting function plotted over $\theta_G[k]$; indeed, intuitively poor grasps are penalized heavily and the metric is minimized at the ideal grasp.  For convenience, we consolidate the variables across the two grasp stability programs for all objects as $\mathbf{u}_B$ and express the total cost and concatenated constraints as $\frac{1}{2}\mathbf{u}_B^\top Q_B(\mathbf{z})\mathbf{u}_B$ and $A_B(\mathbf{z})\mathbf{u}_B\leq b_B(\mathbf{z})$, $H_B(\mathbf{z})\mathbf{u}_B= g_B(\mathbf{z})$, respectively; these matrices are defined in \cref{sec:GSAppendix}.  The total cost over optimized grasp variables $\mathbf{u}_B^*$ is referred to as $J_B^*(\mathbf{z}):=\frac{1}{2}\mathbf{u}_B^{*\top}Q_B(\mathbf{z})\mathbf{u}_B^*$.

\section{Gripper shape}\label{sec:gripperShape}
In this section, we develop a convex QP that, if feasible, yields optimized gripper shapes that contact all objects as specified by the candidate configuration variables $\mathbf{z}$.  We parameterize each gripper's contact surface shape with a piecewise cubic Hermite interpolating polynomial, optimizing over horizontal positions of the left and right gripper surfaces, $v_L[i_y]$, $v_R[i_y]$ and slopes $m_L[i_y]$, $m_R[i_y]$ along a fixed, uniform grid $\mathbf{y}$ of breakpoints along the vertical axis. We consolidate $\mathbf{V}:=[v_L[0],...,v_L[N_y],v_R[0],...,v_R[N_y]]^\top $, and $\mathbf{M}:=[m_L[0],...,m_L[N_y],m_R[0],...,m_R[N_y]]^\top $. 

The gripper must meet each contact $i$ in the gripper frames $G_j$ at the correct location, $\mathbf{p}_i^{G_j}$, with surface tangent parallel to the contacted face, $\mathbf{t}_i^{G_j}$.  We impose these conditions by interpolating the polynomials describing gripper surface location and slope at vertical coordinates corresponding to contact points.

\begin{figure}[t]
    \centering
    \includegraphics[width=\columnwidth]{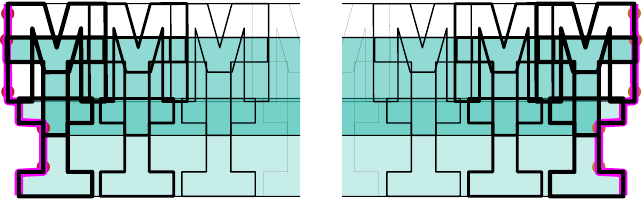}
    \caption{Objects in gripper frames (left $G_L$ and right $G_R$; jaw closing axis is horizontal) along grasp process trajectory for configuration variables $\mathbf{z}$ from \cref{fig:intro3}.  Objects are swept along the horizontal trajectory, with some object outlines along the trajectory traced in black lines, thickening along the trajectory.  To avoid penetrating the objects along this trajectory, the left and right gripper surfaces must lie to the left and right of the magenta lines in their respective frames.} 
    \label{fig:sweep}
    \vspace{-2pt}
\end{figure}

We saw in \cref{sec:approach} that each gripper surface must not intersect any object in the corresponding gripper frame.  A simple extension of this concept allows us to consider the grasping process.  We assume the gripper jaws close linearly on the object from an arbitrarily large separation distance.  The gripper must not intersect the objects in the gripper frame as it moves along this trajectory.  \cref{fig:sweep} shows how this constraint amounts to upper- and lower-bounding the left and right gripper shape horizontal coordinates, respectively.  Optionally, user-specified obstacles can also be included in the non-penetration formulation.  An example is given in \cref{sec:tools}.  The full gripper shape program our framework uses is given in \autoref{eq:gripperShape}.
\begin{subequations}
\begin{align}
&J_S^*(\mathbf{z})  :=  \min_{\mathbf{V,M}} \quad J_S(\mathbf{V,M})&\\
&\textrm{such that} \nonumber\\
& \mathrm{f_c}(\mathbf{z},\mathbf{V},\mathbf{M},i)=[1,0]\mathbf{p}_i^{G_j}&\forall& j, i\in C_j &\label{eq:contact}\\
& \mathrm{f_t}(\mathbf{z},\mathbf{V},\mathbf{M},i)=\frac{[1,0]\mathbf{t}_i^{G_j}}{[0,1]\mathbf{t}_i^{G_j}}&\forall& j, i\in C_j &\label{eq:tangent}\\
  &v_L[i_y]\leq b_U(\Theta,\mathcal{D},i_y) &\forall& i_y \label{eq:nonpenUB}\\
  &v_R[i_y]\geq b_L(\Theta,\mathcal{D},i_y) &\forall& i_y \label{eq:nonpenLB}\\
 &v_L[i_y]-v_R[i_y]\leq \min_k\gamma[k]&\forall& i_y,\label{eq:nonInterpen}
\end{align}\label{eq:gripperShape}
\end{subequations}%
\looseness=-1 \hspace{-5pt}where $J_S[k]$ is a shape quality metric (defined later), $f_c$ and $f_t$, defined in \cref{sec:shapeAppendix}, interpolate the position and derivative of the gripper contact surface, and $b_U(\Theta,\mathcal{D},i_y)$ and $b_L(\Theta,\mathcal{D},i_y)$ give the non-penetration upper and lower bounds of the gripper position coordinates in $G_L$ and $G_R$, respectively. Constraints \autoref{eq:contact} and \autoref{eq:tangent} enforce contact location and tangent, \autoref{eq:nonpenUB} and \autoref{eq:nonpenLB} enforce gripper-object non-penetration, and \autoref{eq:nonInterpen} constrains that the two gripper surfaces are mutually collision-free at their closest approach.  This is achieved by enforcing that the smallest jaw opening (RHS) upper-bounds the overlap that would occur between the left and right gripper surfaces with no jaw opening (LHS).

To encourage shape regularity, especially near contact points, we define a cost term penalizing the sum of squares of second derivatives at the piecewise polynomial breakpoints, Gaussian-weighted by distance from contact points.  To discourage circuitous features, we include an additional term approximating shortest-path. This full cost function, $J_S$, is given in \cref{sec:shapeCostAppendix}.

$J_S$ is a convex quadratic function of $\mathbf{V}$ and $\mathbf{M}$ and can be written in the form $\frac{1}{2}\mathbf{u}_S^\top Q_S(\mathbf{z})\mathbf{u}_S$ where $\mathbf{u}_S:=[\mathbf{V}^\top ,\mathbf{M}^\top ]^\top $. To abbreviate the constraints of \autoref{eq:gripperShape}, let inequalities \autoref{eq:nonpenUB} to \autoref{eq:nonInterpen} be represented by $A_S(\mathbf{z})\mathbf{u}_S\leq b_S(\mathbf{z})$.  Let equalities \autoref{eq:contact} and \autoref{eq:tangent} be represented by $H_{S}(\mathbf{z})\mathbf{u}_S=g_S(\mathbf{z})$.

\section{The full optimization problem}\label{sec:optimization}
As developed in \cref{sec:graspStability,sec:gripperShape}, the feasibility and quality of the gripper resulting from candidate configuration variables $\mathbf{z}$ can be assessed via two convex QPs whose parameter matrices are functions of $\mathbf{z}$.  In addition, we include upper and lower bounds on elements of $\mathbf{z}$, denoted $\mathbf{z}\in Z$, noting that $Z$ is simply a box.  The resulting full NLP is \autoref{eq:fullNLP}:
\begin{equation}\label{eq:fullNLP}
    \mathbf{z}^*:= \argmin_{\mathbf{z}\in Z} \left(J^*_B(\mathbf{z}) + J^*_S(\mathbf{z})\right)
\end{equation}
The minimization in \autoref{eq:fullNLP} cannot be solved directly via NLP solvers because in significant subsets of $Z$, \autoref{eq:graspStabilityW} and \autoref{eq:gripperShape} are infeasible, making $J^*_B$ and $J^*_S$ undefined. Instead we use an augmented Lagrangian formulation, where objective values can still be evaluated when QP constraints are violated.

To represent all constraints as equalities, we introduce slack variables $\mathbf{s}_B,\mathbf{s}_S\geq 0$.  We consolidate all grasp stability and shape variables as $\mathbf{u}:=[\mathbf{u}_B^\top ,\mathbf{s}_B^\top ,\mathbf{u}_S^\top ,\mathbf{s}_S^\top ]^\top $ such that $A(\mathbf{z})\mathbf{u}=b(\mathbf{z})$, where
$$A(\mathbf{z}):=\begin{bmatrix}A_B(\mathbf{z}) &I&0&0\\ H_B (\mathbf{z})&0&0&0\\0&0&\rho_SA_S(\mathbf{z})&I\\0&0&\rho_SH_S(\mathbf{z})&0\end{bmatrix},\ b(\mathbf{z}):=\begin{bmatrix}
    b_B(\mathbf{z})\\g_B(\mathbf{z})\\\rho_Sb_S(\mathbf{z})\\\rho_Sg_S(\mathbf{z})
\end{bmatrix},$$
and $\rho_S$ weights the shape constraints. We leave the bounds $\mathbf{z}\in Z$ and $\mathbf{u}\in U:=\{\mathbf{u}|\mathbf{s}_B,\mathbf{s}_S\geq 0 \}$ as hard constraints and form the (partially) augmented Lagrangian:
\begin{equation}\label{eq:augmentedLagrangian}
\begin{split}
    L(\mathbf{z},\mathbf{u},\nu) := \frac{1}{2}\mathbf{u}_B^\top Q_B(\mathbf{z})\mathbf{u}_B +\frac{w}{2}\mathbf{u}_S^\top Q_S(\mathbf{z})\mathbf{u}_S +\\
    \nu^\top (A(\mathbf{z})\mathbf{u}-b(\mathbf{z})) + \frac{\rho}{2}||A(\mathbf{z})\mathbf{u}-b(\mathbf{z})||_2^2, 
    \end{split}
\end{equation}
where $\nu$ are Lagrange multipliers. To take minimizing steps over the primal variables $\mathbf{z},\mathbf{u}$, note that the $\mathbf{z}^{L*}$ calculated with  $(\mathbf{z}^{L*},\mathbf{u}^{L*})=\argmin_{\mathbf{z}\in Z,\mathbf{u}\in U}L(\mathbf{z},\mathbf{u},\nu)$ is equivalent to $\mathbf{z}^{L*}=\argmin_{\mathbf{z}\in Z}L^*(\mathbf{z},\nu)$ where $L^*(\mathbf{z},\nu):=\min_{\mathbf{u}\in U}L(\mathbf{z},\mathbf{u},\nu)$.  While minimizing over $\mathbf{z}$ is nonconvex, the inner minimization $\min_{\mathbf{u}\in U}L(\mathbf{z},\mathbf{u},\nu)$ can be done globally and efficiently, as it is a convex QP for fixed $\mathbf{z}$.
We solve iteratively, taking steps minimizing over $\mathbf{z}$ with an NLP solver initialized at the current $\mathbf{z}$ solution, and updating $\nu^+\leftarrow \nu + \rho(A(\mathbf{z})\mathbf{u}-b(\mathbf{z}))$ and $\rho^+\leftarrow\phi\rho$.  

This formulation is nonconvex with many local minima.  As a mitigation, before every $\mathbf{z}$ update we check whether the $\mathbf{z}$ solution from any previous iteration results in a lower $L^*(\mathbf{z},\nu)$ than the current $\mathbf{z}$ solution, restoring the previous solution if so. 
Again due to nonconvexity, we uniformly randomize the initial guess for the vertical components of gripper positions and solve using multiple randomized initial guesses. We randomize this particular parameter because reordering objects vertically in the gripper frames requires the optimizer to pass through high-cost ($L^*(\mathbf{z},\nu)$) candidate configurations that have penetration. This discourages full exploration of the space of gripper position vertical coordinates.

\subsection{Post processing}\label{sec:postProcessing}
The augmented Lagrangian method never enforces hard constraints and eventually incurs a trade-off between numeric stability and constraint satisfaction as $\rho$ grows over iterations.  In addition, even when constraints in \autoref{eq:gripperShape} are satisfied, the finite discretization in $\mathbf{y}$ can leave small penetrations between the interpolated gripper surfaces and the objects.  As such, we send the best solutions through two post-processing steps and discard the rest.

\begin{figure}[t]
    \centering
    \includegraphics[width=\columnwidth]{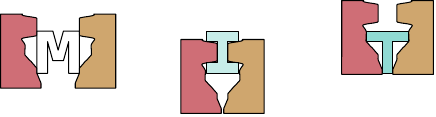}
    \caption{The second-highest-ranking local minimum solution for grasping letters.  Unlike in \cref{fig:intro4}, the gripper jaw opening distance varies significantly across grasps.} 
    \label{fig:lettersTopDifferentSoln}
    \vspace{-2pt}
\end{figure}

\begin{figure}[t]
    \centering
    \includegraphics[width=\columnwidth]
    {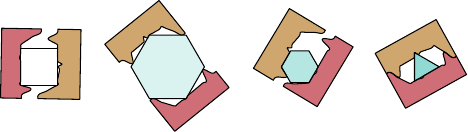}
\includegraphics[width=0.89\columnwidth]{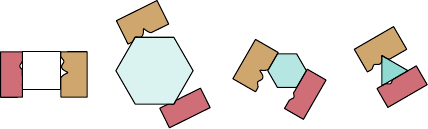}
    
    \caption{Solutions for grasping a set of polygons. Top: The highest-ranking solution. Bottom: Another high-ranking solution.} 
    \label{fig:polygonsTopSoln}
    \vspace{-2pt}
\end{figure}

  \begin{figure*}[t]
  \sbox\twosubbox{%
  \resizebox{\dimexpr \linewidth-1em}{!}{%
    \includegraphics[height=1cm]{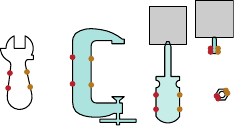}\vspace{5pt}
    \hspace{1cm}
    \includegraphics[height=0.85cm]{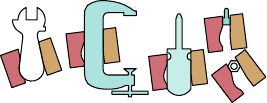}%
  }%
}
\setlength{\twosubht}{\ht\twosubbox}
\centering
\subcaptionbox{ \label{fig:toolsDef}}{%
  \includegraphics[height=\twosubht]{figures/toolsDef_2.pdf}\vspace{5pt}
}\hspace{1cm}
\subcaptionbox{\label{fig:toolsSoln}}{%
  \includegraphics[height=0.85\twosubht]{figures/toolsSoln_2_cropped.pdf}%
}

  \sbox\twosubbox{%
  \resizebox{\dimexpr \linewidth-1em}{!}{%
\includegraphics[height=1cm]{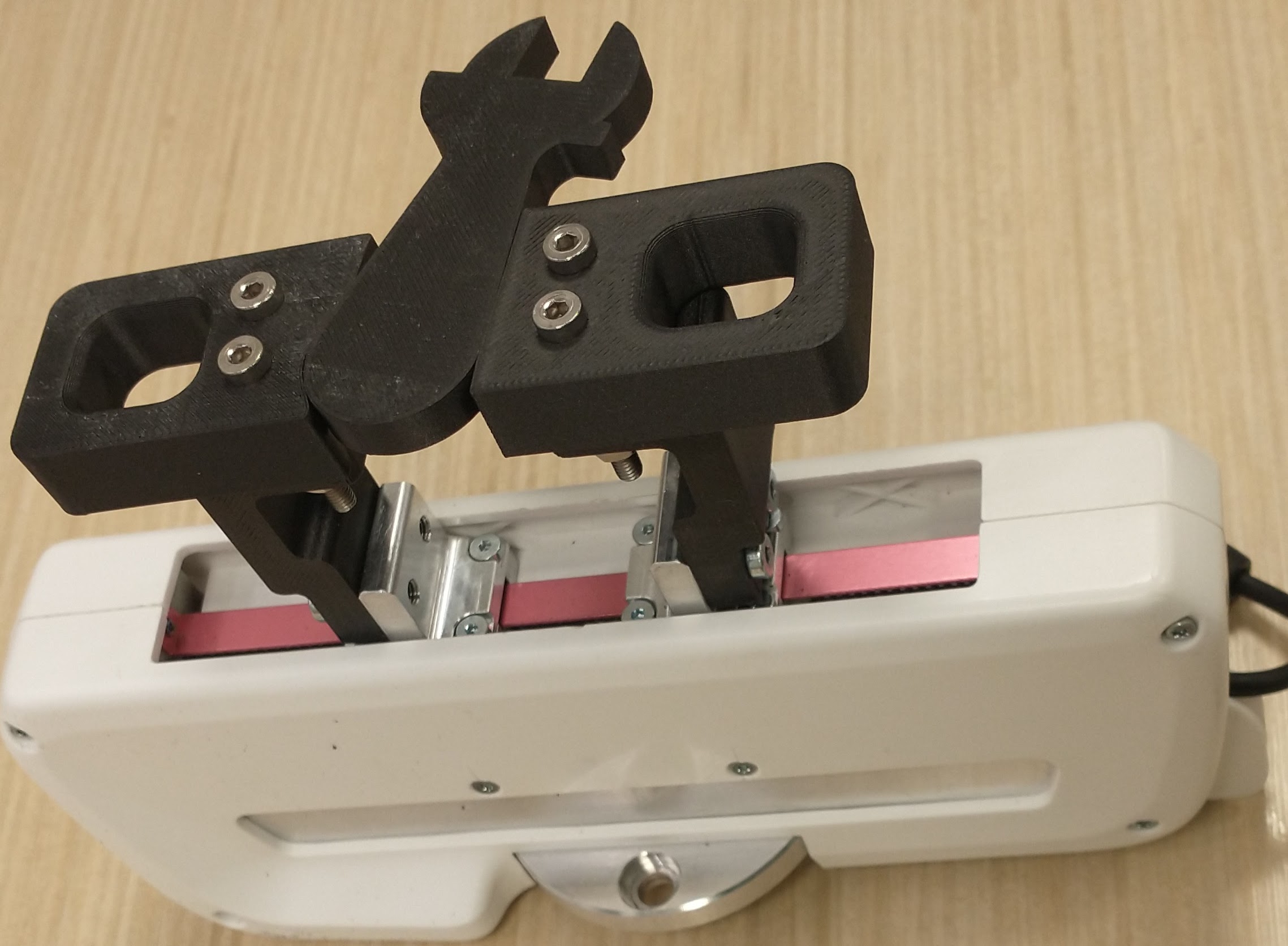}
\quad
    \includegraphics[height=1cm]{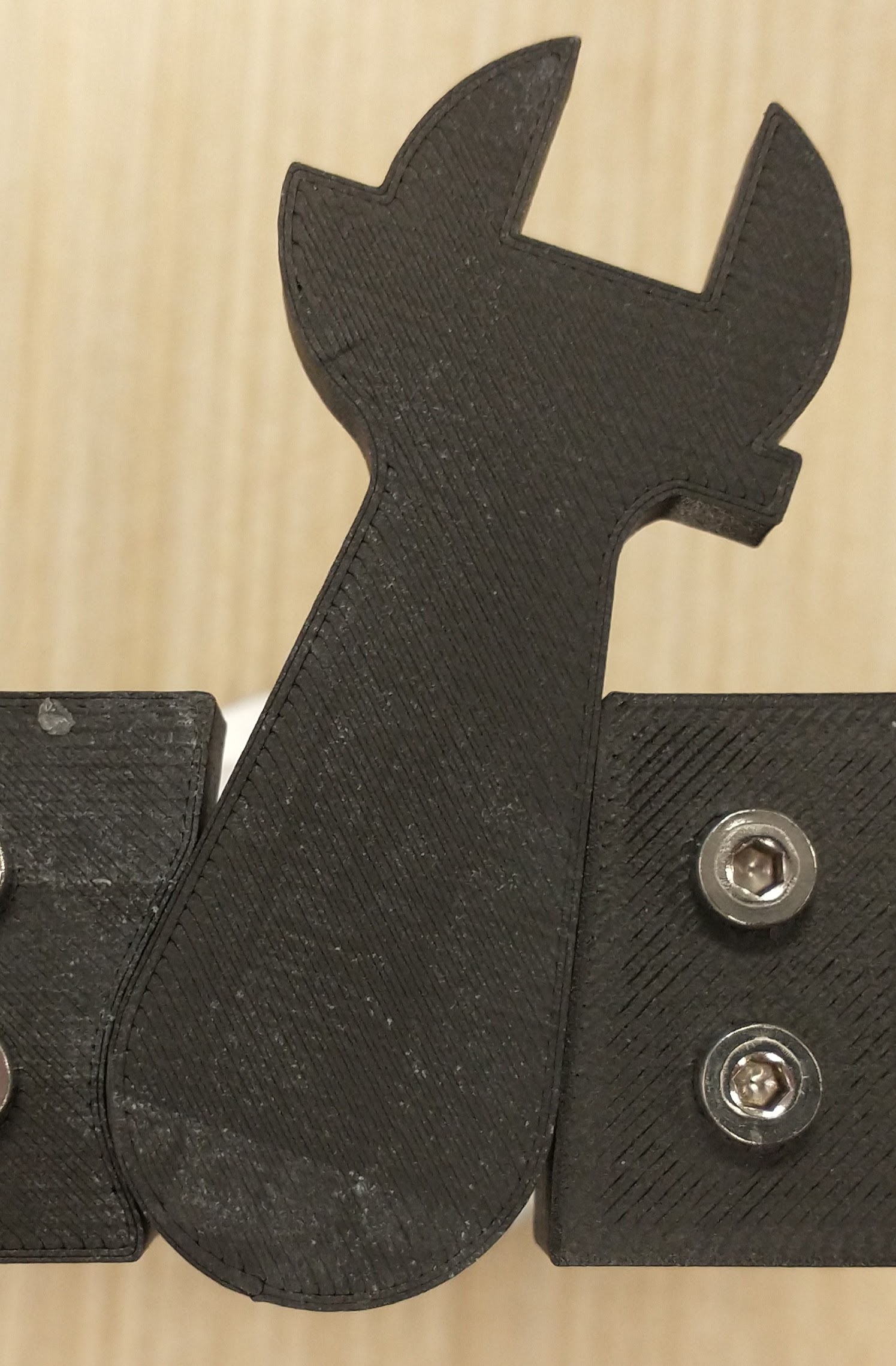}%
    \includegraphics[height=1cm]{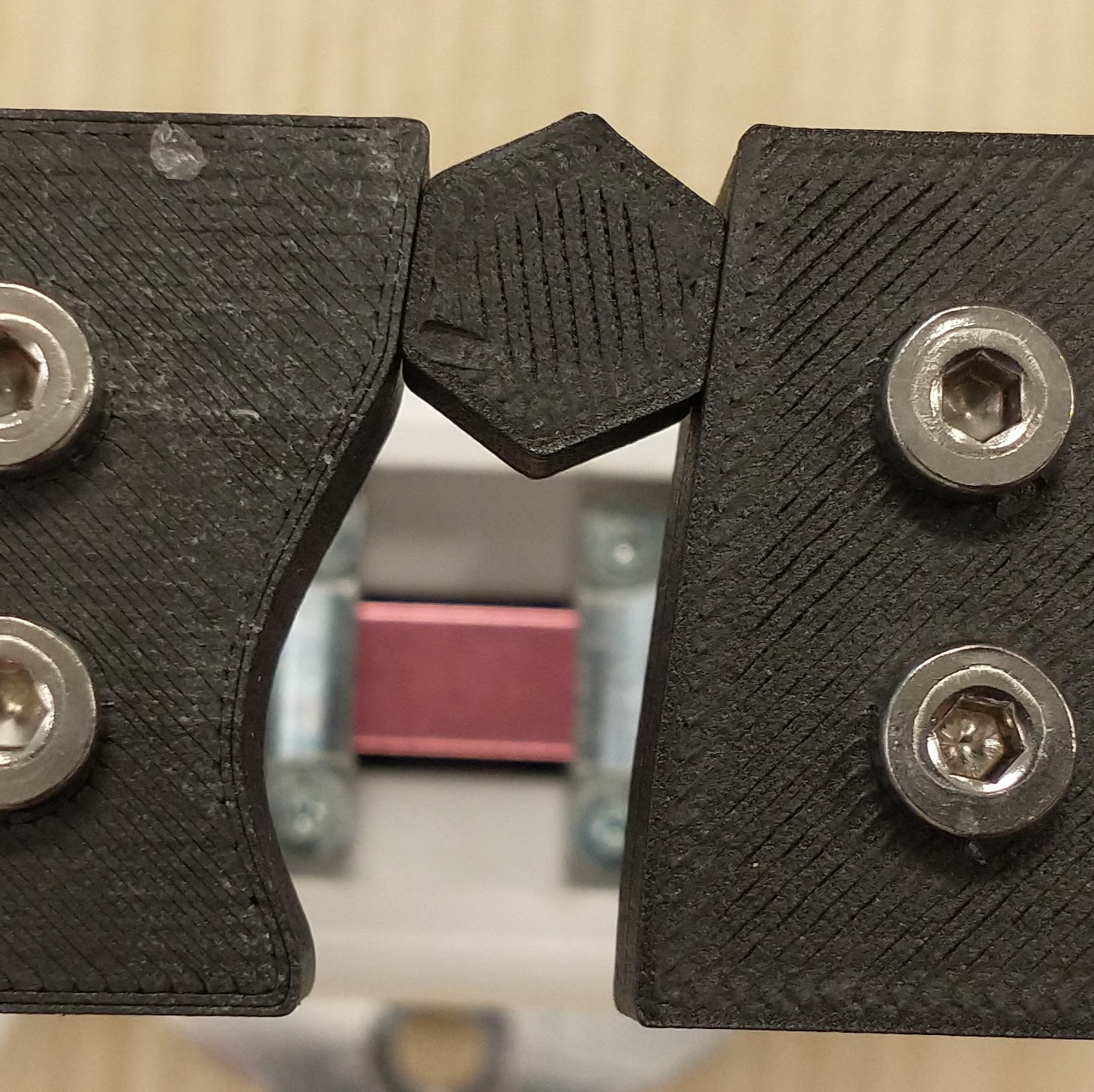}%
    \includegraphics[height=1cm]{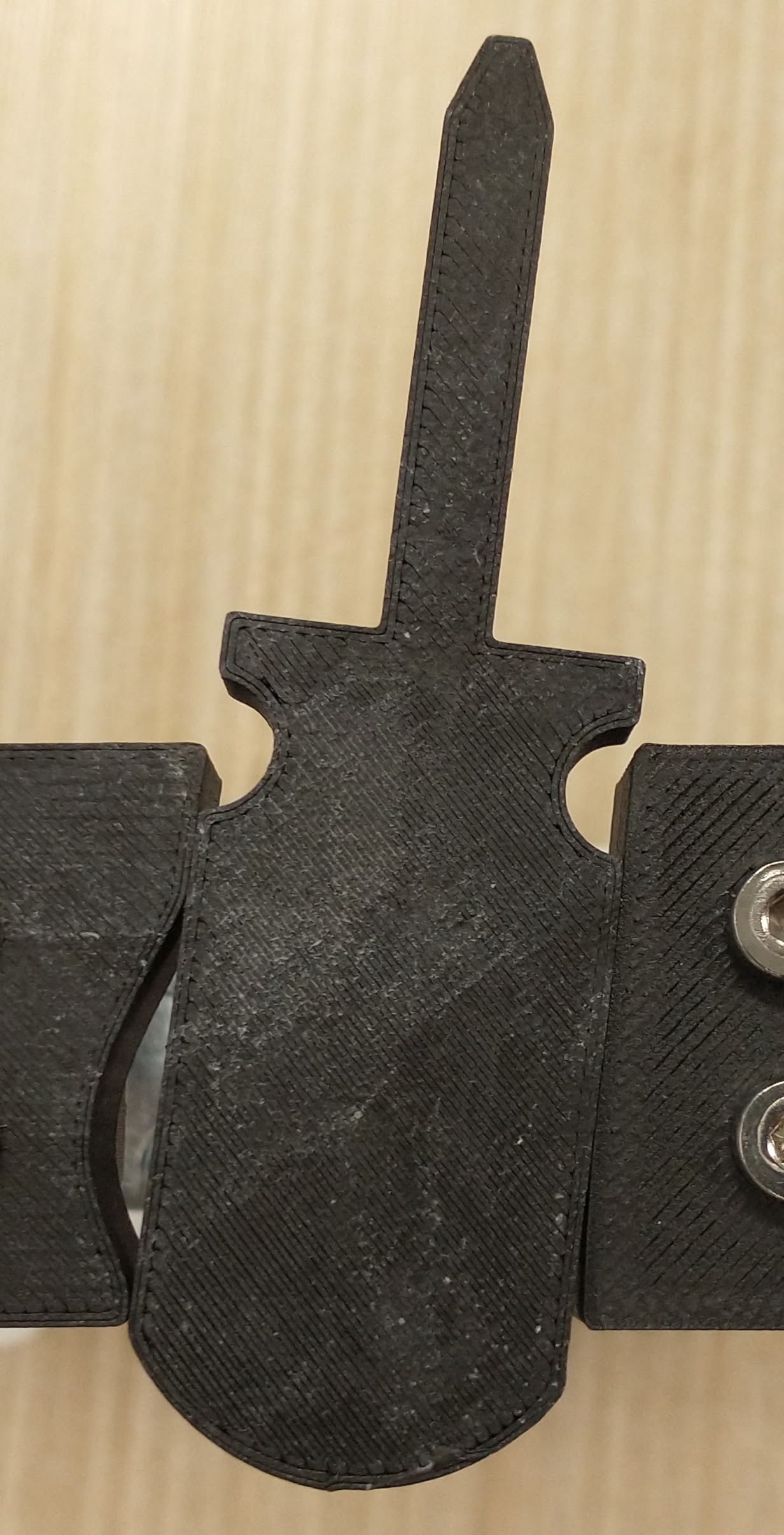}%
    \includegraphics[height=1cm]{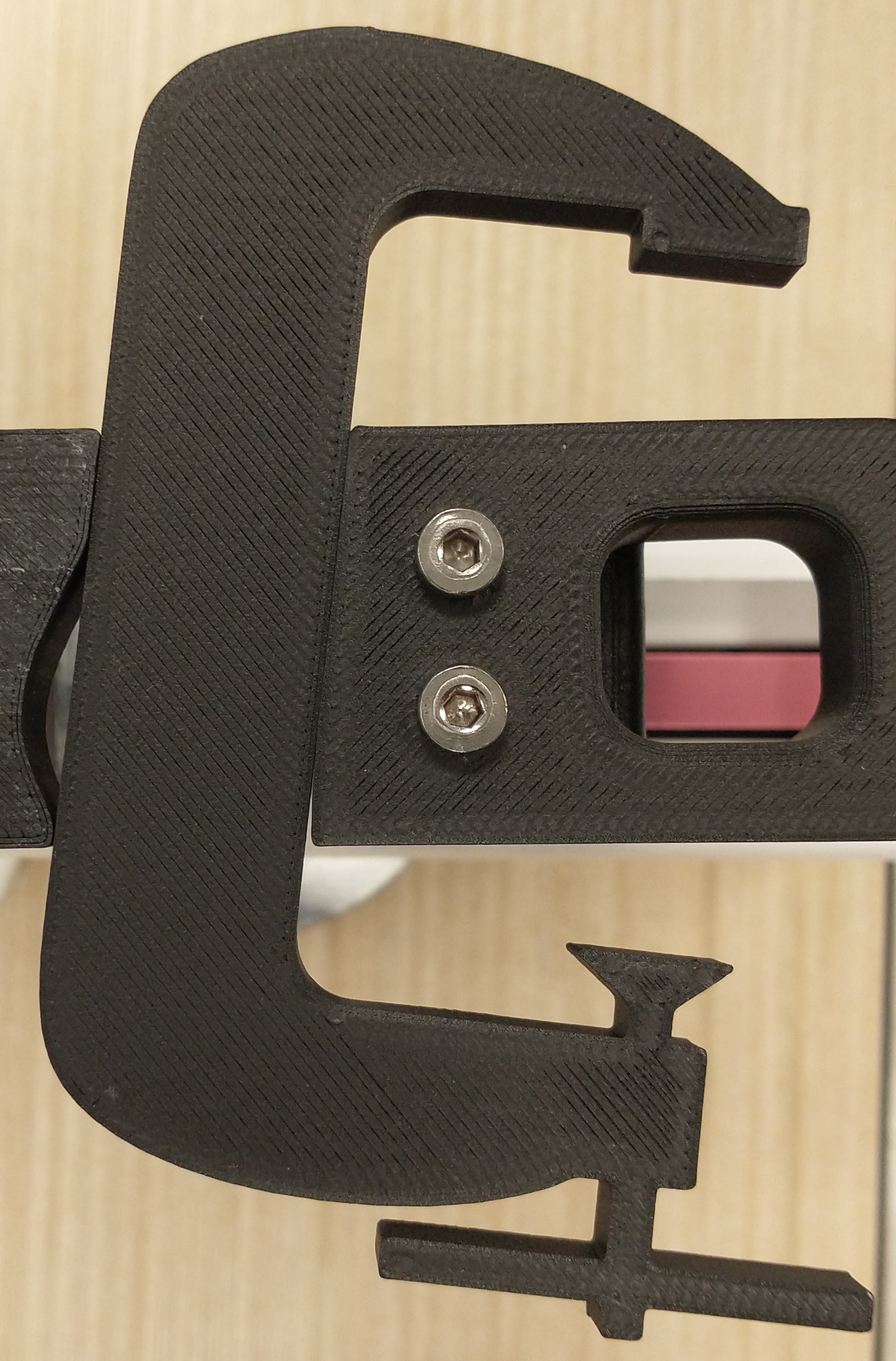}%
    \includegraphics[height=1cm]{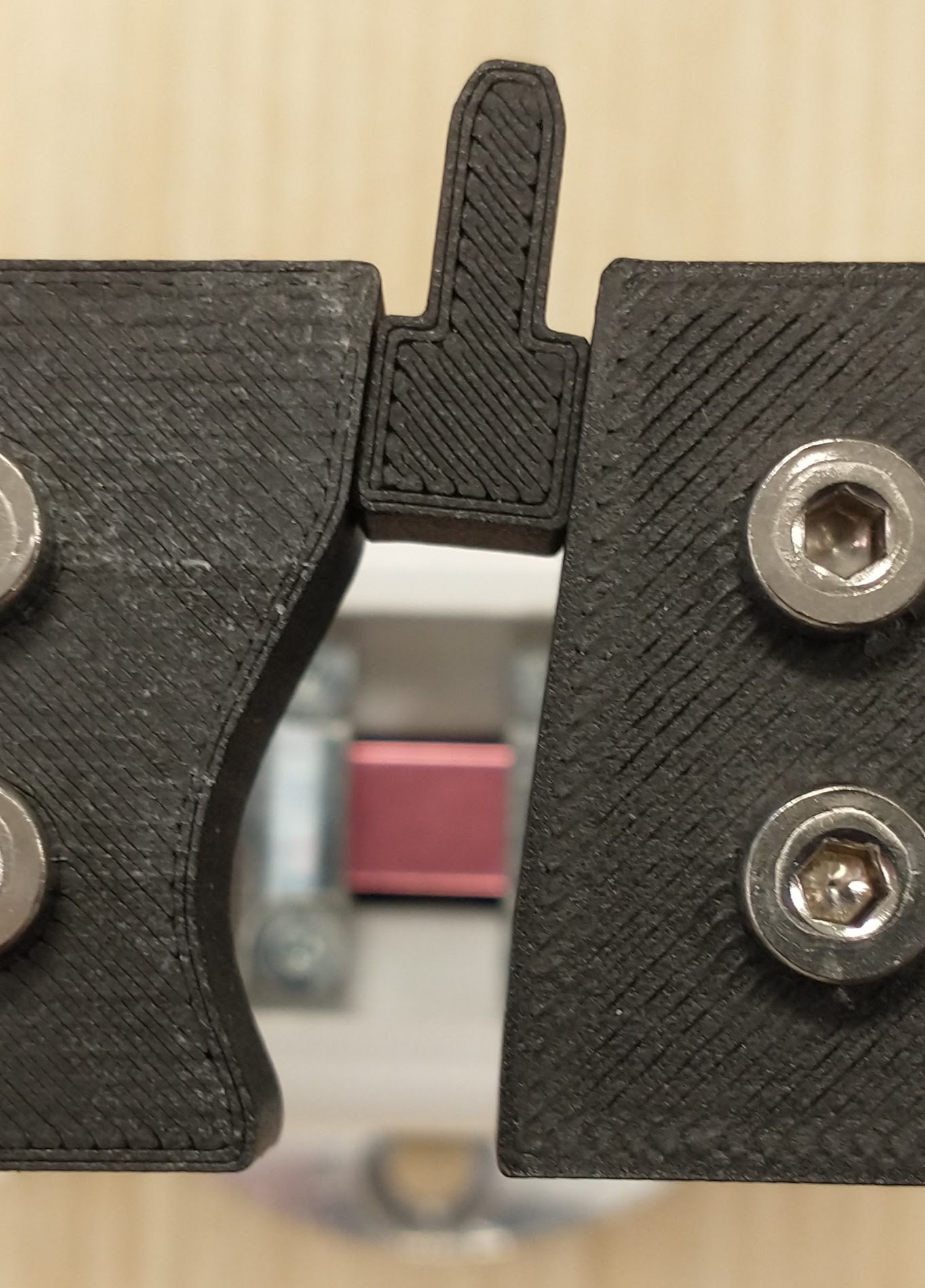}%

  }%
}
\setlength{\twosubht}{\ht\twosubbox}

\subcaptionbox{\label{fig:demoOut}}{%
\includegraphics[height=\twosubht]{figures/zoomedOutDemo.jpg}}
\quad
\subcaptionbox{\label{fig:demoIn}}{%
    \includegraphics[height=\twosubht]{figures/wrench.jpg}%
    \includegraphics[height=\twosubht]{figures/nut.jpg}%
    \includegraphics[height=\twosubht]{figures/screwdriver.jpg}%
    \includegraphics[height=\twosubht]{figures/clamp.jpg}%
    \includegraphics[height=\twosubht]{figures/screw.jpg}%
}
\caption{(a) Input information for grasping a wrench, C-clamp, screwdriver, screw, and nut.  The screwdriver and screw descriptions include obstacles (gray) to ensure that the gripper solution does not geometrically conflict with a workpiece. (b) highest-ranked solution to toolset-grasping problem. (c-d) Hardware demonstration of these grasps.  (c) shows the Franka Emika Robot Hand, mounted with the optimized gripper, actuated manually for this demonstration.}
\label{fig:toolsFig}
\vspace{-18pt}
\end{figure*}

\textbf{Stage A} makes small adjustments to configuration variables solution $\mathbf{z}$ to ensure that all contacts are possible to access.  Recall from \cref{sec:gripperShape} that the left and right gripper surface horizontal coordinates are upper and lower bounded by the object boundaries in the gripper frames.  Thus, if a contact point resides beyond these bounds, the gripper surface cannot possibly meet it.  Even reasonable-cost ($L$) outputs of the main optimization phase often slightly violate this condition due to soft constraints.  

As such, we use a secondary optimization process similar to the first but with a few modifications. First, we hard-enforce signed-distance function constraints on the contact points $\mathbf{p}_i^{G_j}$ relative to the objects $\Psi^{G_j}[k]$ in order to ensure that contact can be made without penetration. Second we restrict configuration variables to be near the original output via bounds $\mathbf{z}\in Z'$. Third, we restrict the span of grid $\mathbf{y}$ to the relevant span of vertical coordinates near contact points in the original output.  Finally, in this phase, we hold the penalty parameter $\rho$ constant at its final value from the main optimization phase and set Lagrange multipliers $\nu=0$, making \cref{eq:augmentedLagrangian} the quadratic penalty function. If a solution is found, stage A outputs an updated $\mathbf{z}$ near the solution from the main optimization phase, such that contact points $\mathbf{p}_i^{G_j}$ are all outside objects $\Psi^{G_j}[k]$ in the gripper frames.

\textbf{Stage B} solves for the gripper shape $\mathbf{u}_S$ via \autoref{eq:gripperShape}.  The span of grid $\mathbf{y}$ is restricted to the relevant span of vertical coordinates near contact points, and breakpoints are added at the vertical coordinates of contact points and object vertices.  This achieves exact satisfaction of contact and non-penetration constraints at these points.

\section{Results}\label{sec:results}

Here we present results from example problems.  We continue discussing the letter-grasping problem, and introduce two new object sets: polygons and tools.
\subsection{MIT letters}

The solution shown in \cref{fig:intro4} dominated the top solutions for the letter-grasping problem defined in \cref{fig:intro1}, resulting from many initial guesses, which strengthens confidence in this solution's optimality.   The solution is simple, aligning contact points to be contacted with the same gripper features. \cref{fig:lettersTopDifferentSoln} shows the
next-highest-ranking solution that is qualitatively different.  Both solutions maintain the letters in the orientations that globally optimize grasp stability, and keep contact points spread maximally far apart on each edge, but the solution in \cref{fig:lettersTopDifferentSoln} does not re-purpose features as well and thus has a more complex shape with higher cost.

\subsection{Polygons}
We solve a polygon-grasping problem with more restrictive limits on the vertical gripper positions, simulating scenarios where grippers must be found with restricted dimensions. \cref{fig:polygonsTopSoln} (top) shows the top-ranking solution.    The second- and third-ranking solutions are similar, but, due to the up-down symmetry of this set of objects, flipped without consequence.  The large hexagon is tilted slightly off its grasp-stability-optimal orientation because, if level, the features contacting it on either jaw would interpenetrate during the triangle grasp.  Another high-ranking solution with very simple gripper shapes is shown in \cref{fig:polygonsTopSoln} (bottom).

\subsection{Toolset}\label{sec:tools}

Suppose an assembly task requires stably grasping a variety of tools and fasteners.  Furthermore, suppose the intended task requires that the gripper clear a workpiece, for example a surface that a screw is installed into.  Such a problem input is shown in \cref{fig:toolsDef}, including obstacles associated with the screwdriver and screw to represent the workpiece.  The slanted faces of the wrench handle prevent use of flat parallel jaws, and contacting in the concavity of the C-clamp restricts the gripper vertically on one side.  Additionally, these objects have many faces and nonconvexities, which could pose computational challenges for a model-based approach like ours.  Nonetheless, the solver finds an elegant solution, shown in \cref{fig:toolsSoln}, which repurposes flat faces for many grasps and respects the aforementioned considerations.
We provide a demonstration, shown in \cref{fig:demoIn,fig:demoOut}, of 3D-printed models of the toolset objects being grasped by the 3D-printed optimized grippers.

\section{Parameters and computation}\label{sec:params}
\begin{table*}[t]
\centering
\begin{tabular}{c@{\hskip 0.2in}c@{\hskip 0.2in}c@{\hskip 0.2in}c@{\hskip 0.2in}c@{\hskip 0.2in}c@{\hskip 0.2in}c@{\hskip 0.2in}c@{\hskip 0.2in}c@{\hskip 0.2in}c@{\hskip 0.2in}c@{\hskip 0.2in}c}
    Problem & $N_{IG}$ & $T_{IG} (s)$ & $N_{PP}$ & $T_{PP}(s)$ & $\mathbf{p}_G^W$ bounds &  $N_y$ & $y[0]$ & $y[N_y]$ & $w_s$ & $w_p$ & $L_{max}$ \\ \hline

    Letters & 60 & 280 & 0 &  & $\pm [1,0.5]^\top $ & 50& -1.2 & 1.2 & 3e-4& 0.1 & 0.71\\\ 
     
      Polygons & 60 & 619 & 5 & 81 & $\pm [1,0.3]^\top $ & 50& -1.3 & 1.3 & 3e-4& 0.1 & 1.00\\ 
      
      Toolset & 200 & 1454 & 5 & 190 & $\pm [1,0.8]^\top $ & 100 & -2.2 & 2.2&3&0.01 &1.12 \\ 
\end{tabular}
\caption{Computational parameters for the example problems in \cref{sec:results}.  Here, $N_{IG}$ is the number of randomized initial guesses, $T_{IG}$ is the average solve time per initial guess, $N_{PP}$ is the number of solutions sent to stage A post-processing, $T_{PP}$ is the average solve time per stage-A post-processing run, and $L_{max}$ is the maximum length $\max_kL[k]$.  The letters problem was simple and well-behaved enough that stage A post-processing was not needed.}
\vspace{-16pt}
\label{tab:param}
\end{table*}
 We share problem parameters in \cref{tab:param}.  In addition to these parameters, the following are constant across all experiments:
\begin{itemize}[leftmargin=*,wide]
\item $\mu = 0.3$, $\gamma[k]\in [-1, 4]\ \forall k$, $\phi = 2$, $\rho_S = 3$, $\sigma = 0.2$.
\item $\theta_G$ bounds are calculated by first finding a value where \autoref{eq:graspStabilityW} is feasible for $w=[0,0,0]^\top $, then finding the first point in each direction where either this program is infeasible or a contact tangent becomes horizontal (impossible for the gripper shape to meet with finite slope).  Thus, the algorithm is invariant to input object orientation in $W$.
    \item We bound the contact points between 0.1 and 0.9 of the extent of the contacted object face, because the grasp stability QP models contact faces as infinite and cannot consider robustness consequences of contacting close to vertices.
    \item We run the main optimization phase for 30 iterations.
    \item We solve the augmented Lagrangian minimization over $\mathbf{z}$ with SNOPT  \cite{snopt77,GilMS05}.
\end{itemize}

Our problem scales well: The number of configuration variables $\mathbf{z}$ is linear in the number of objects.  All presented problems are solved on a computer with Intel Core i7-10750H 2.60GHz CPU. As is intuitive according to object number and shape complexity, the letters-grasping problem is solved fastest, and the toolset-grasping problem is solved slowest.  Gripper design is always performed offline, and is not expected to be fast. Consequently, we have not focused on computational efficiency, and our implementation could be further optimized for speed.  

The discretization of the shape problem \autoref{eq:gripperShape} presents issues for gradients.  Namely, as object features and contact points move vertically, dependencies of cost and constraint components on the optimization variables $\mathbf{u}_S$ change discretely.  For example, as the lowest object in the gripper frames moves upward along the $\mathbf{y}$ grid, the number of points $v_L[i_y]$ and $v_R[i_y]$ that have finite bounds due to non-penetration decreases. Future approaches could involve gradient smoothing, or using NLP solvers that incorporate momentum.

\section{Conclusions}\label{sec:conclusions}
We propose an NLP framework for optimizing parallel-jaw grippers for grasping sets of objects, given suitable contact point assignments to object faces and gripper jaws.  We use a high-dimensional, expressive parameterization for gripper shape, and leverage a division of problem variables to make the problem tractable without reducing the design space.  We constrain feasibility and optimize the quality of the grasps resulting from these grippers, additionally including metrics for shape regularity.  This formulation yields solutions to several example problems, which validate that the framework can produce optimal, simple solutions where possible, find grippers of constrained dimensions, and handle obstacles and complex objects.

This work assumes accurate models.  
While the proposed framework does prevent geometric conflicts in the modeled grasping process (\cref{fig:sweep}), shape or pose variations could cause the object to escape the grasp as the jaws close, or could lead to reduced grasp stability.  
A user could employ the current framework to generate more robust solutions by providing, for each object, a set of input objects whose geometries and poses are variations of the expected values, and constraining that the gripper pose solution is the same for each.  This is analogous to the use of randomized object poses used in aforementioned simulation-based frameworks \cite{wolniakowski2015task,aletrobotic,Ha2020fit2form}. In contrast, grasp stability and non-penetration for each input object are hard-constrained in our framework, which creates a trade-off: an output solution is guaranteed to meet constraints, but the problem can become infeasible depending on input specification.

In the future, this framework could be extended to optimize over the number of contacts, and assignments of those contacts to object faces and gripper jaws.  While this problem is discrete by nature, it may have a tractable and informative relaxation.  Furthermore, future work could consider the three-dimensional problem, which may lend itself to more interesting geometries and approaches to stabilizing grasps. Another item of future work is to investigate more elegant mitigations for nonconvexity and nonsmoothness.   

Finally, concepts from this implementation may be fruitfully borrowed for manipulator-design problems with more degrees of freedom and trajectories.  In particular, both in this work and our previous work on co-optimizing shape and motion of rigid effectors \cite{jiang2022shape}, representing objects, obstacles, and contact points in frames fixed to rigid manipulator links facilitated natural expression of constraints on the manipulator's shape, as well as tractable scaling of the design space.

\appendices
\vspace{-6pt}
\crefalias{section}{appendix}
\section{Grasp stability program matrix consolidation}
\label{sec:GSAppendix}
\vspace{-5pt}
As \autoref{eq:graspStabilityW} is evaluated twice for each object, there are $2(N_o+1)$ instances of variables $\mathbf{r}, \mathbf{q},\mathbf{c}$.  Let $\mathbf{u}_{B+}:=[\mathbf{r}[0]_+^\top ,\mathbf{q}[0]_+^\top ,\mathbf{c}[0]_+^\top ,...,\mathbf{r}[N_o]_+^\top ,\mathbf{q}[N_o]_+^\top ,\mathbf{c}[N_o]_+^\top ]^\top $ give the variables corresponding to \autoref{eq:graspStabilityW} with $\mathbf{w}=\mathbf{w}_+:=[0,0,1]^\top $, and $\mathbf{u}_{B-}$ be defined similarly for $\mathbf{w}=\mathbf{w}_-:=[0,0,-1]^\top $, and $\mathbf{u}_B:=[\mathbf{u}_{B+}^\top ,\mathbf{u}_{B-}^\top ]^\top $.
The cost function used to judge the grasp stability of all grasps in the object set is:
\begin{equation*}\label{eq:fullGraspStabilityCost}\small
\begin{split}
   J_B(\mathbf{z}):= \sum_{k=0}^{N_o} \left(\tilde{J}_{B}[k](\mathbf{r}_+[k],\mathbf{q}_+[k])+\tilde{J}_{B}[k](\mathbf{r}_-[k],\mathbf{q}_-[k])\right)\\
   =\sum_{k=0}^{N_o} \tilde{J}^*_B[k](\mathbf{z},\mathbf{w}_+)+\tilde{J}^*_B[k](\mathbf{z},\mathbf{w}_-)
   =\frac{1}{2}\mathbf{u}_B^\top Q_B(\mathbf{z})\mathbf{u}_B
    \end{split}
\end{equation*}
where we define $Q_B$ noting that the cost is quadratic in $\mathbf{u}_B$.  To consolidate the constraint matrices across the $2(N_o+1)$ instances of \autoref{eq:graspStabilityW}, note that only the RHS of the equality constraints is affected by $\mathbf{w}$, and let inequalities \autoref{eq:gwSpringNonneg} to \autoref{eq:gFriction} be represented by $$A_{Bk}'[k](\theta_G[k],\mathbf{d}[k])[\mathbf{r}^\top ,\mathbf{q}^\top ,\mathbf{c}^\top ]^\top \leq b_{Bk}'[k](\theta_G[k],\mathbf{d}[k]),$$ 
and let equalities \autoref{eq:gwEquilibrium} and \autoref{eq:gwSpring} be represented by 
$$H_{Bk}'[k](\theta_G[k],\mathbf{d}[k])[\mathbf{r}^\top ,\mathbf{q}^\top ,\mathbf{c}^\top ]^\top =g_{Bk+}'[k](\theta_G[k],\mathbf{d}[k])$$ or 
$$H_{Bk}'[k](\theta_G[k],\mathbf{d}[k])[\mathbf{r}^\top ,\mathbf{q}^\top ,\mathbf{c}^\top ]^\top =g_{Bk-}'[k](\theta_G[k],\mathbf{d}[k])$$ 
for $\mathbf{w} = \mathbf{w}_+ $ or $\mathbf{w} =\mathbf{w}_- $ respectively.  Then let 
$$A_B'(\mathbf{z}):=\textrm{blkdiag}(\{A_{Bk}'[k](\theta_{g}[k],\mathbf{d}[k])\}_{k=0}^{N_o})$$
where $\textrm{blkdiag}(\{X_i\}_{i=0}^N)$ creates a block-diagonal matrix of matrices $\{X_i|i=0,...,N\}$, and $\textrm{blkdiag}(X,Y)$ creates a block-diagonal matrix of the two matrices $X$ and $Y$.  Let
$$\resizebox{\hsize}{!}{$b'_{B}(\mathbf{z}):=[b_{Bk}'[0](\theta_G[0],\mathbf{d}[0])^\top ,...,b_{Bk}'[N_o](\theta_G[N_o],\mathbf{d}[N_o])^\top ]^\top ,$}$$ 
$$H_B'[k](\mathbf{z}):=\mathrm{blkdiag}(\{H_{Bk}'[k](\theta_{G}[k],\mathbf{d}[k])\}_{k=0}^{N_o}),$$ 
$$\resizebox{\hsize}{!}{$g'_{B\pm}(\mathbf{z}):=[g_{Bk\pm}'[0](\theta_G[0],\mathbf{d}[0])^\top ,...,g_{Bk\pm}'[N_o](\theta_G[N_o],\mathbf{d}[N_o])^\top ]^\top .$}  $$
Finally, let $A_B(\mathbf{z}):=\mathrm{blkdiag}(A_B'(\mathbf{z}),A_B'(\mathbf{z}))$, $b_B(\mathbf{z}):=[b'_B(\mathbf{z})^\top ,b'_B(\mathbf{z})^\top ]^\top $, $H_B(\mathbf{z}):=\mathrm{blkdiag}(H_B'(\mathbf{z}),H_B'(\mathbf{z}))$, and $g_B:=[g'_{B+}(\mathbf{z})^\top ,g'_{B-}(\mathbf{z})^\top ]^\top $ such that $A_B(\mathbf{z})\mathbf{u}_B\leq b_B(\mathbf{z})$ and $H_B(\mathbf{z})\mathbf{u}_B= g_B(\mathbf{z})$.

\vspace{-1pt}
\section{Gripper shape interpolation functions}\label{sec:shapeAppendix}
\vspace{-1pt}
Let $f_c$ be a function that, for contact $i$ with jaw assignment $j\in \{L,R\}$, finds vertical grid interval index $i_y$ such that 
$$y[i_y]<[0,1]\mathbf{p}^{G_j}_i\leq y[i_y+1],$$ and calculates
$$l=\frac{[0,1]\mathbf{p}^{G_j}_i-y[i_y]}{y[i_y+1]-y[i_y]},$$ and evaluates to 
\begin{dmath*}
(2l^3-3l^2+1)v_j[i_y]+(-2t^3+3t^2)v_j[i_y+1] + (y[i_y+1]-y[i_y])((t^3-2t^2+t)m_j[i_y] + (t^3-t^2)m_j[i_y+1]).
\end{dmath*}
Similarly, $f_t$ evaluates to 
\begin{dmath*}
    \frac{(6l^2-6l)(v_j[i_y] - v_j[i_y+1])}{y[i_y+1]-y[i_y]} + (3l^2-4l+1)m_j[i_y] + (3l^2-2l)m_j[i_y+1].
\end{dmath*}

\section{Gripper shape cost function}\label{sec:shapeCostAppendix}
Let $f_{t0}$ and $f_{t1}$ be the second derivatives of the piecewise polynomial at the beginning and end of the $i_y$th interval:
\begin{equation*}
    f_{t0}[i_y] := \frac{6(v_j[i_{y+1}]-v_j[i_y])}{(y[i_y+1]-y[i_y])^2}-\frac{2(2m_j[i_y]+m_j[i_{y+1}])}{y[i_y+1]-y[i_y]},
\end{equation*}
\begin{equation*}
    f_{t1}[i_y] := \frac{6(v_j[i_y]-v_j[{y+1}])}{(y[i_y+1]-y[i_y])^2}+\frac{2(m_j[i_y]+2m_j[i_{y+1}])}{y[i_y+1]-y[i_y]}.
\end{equation*}
Then,
\begin{equation*}
    \begin{split}
        J_S(\mathbf{V},\mathbf{M}) = \sum_{j\in \{L,R\}} \sum_{i_y=0}^{N_y-1}\left(w_p'\sum_{i_c\in C_j}\right.(h[i_y,i_c]f_{t0}[i_y]^2+\\
        h[i_y+1,i_c]f_{t1}[i_y]^2) + w_s'(v_j[i_y+1]-v_j[i_y])^2\left.\vphantom{\sum_{i_c\in C_j}} \right),
    \end{split}
\end{equation*}
where 
$h[i_y,i_c] := \exp\left(-(y[i_y]-[0,1]\mathbf{p}^{G_j}_{i_c})^2/(2\sigma^2)\right).$
$w_p'$ and $w_s'$ are scaled versions of input parameters to reduce dependence on $\mathbf{y}$ discretization and object scale: $$w_p':=\frac{w_p\left(\sum_{k=0}^{N_o}L[k]\right)^2}{N_y},w_s':=\frac{w_pN_y^2}{\left(\sum_{k=0}^{N_o}L[k]\right)^2}.$$






\FloatBarrier
\bibliographystyle{IEEEtran} 
{\footnotesize \bibliography{icra_ref}} 

\end{document}